\documentclass[conference]{IEEEtran}
\IEEEoverridecommandlockouts
\usepackage{cite}
\usepackage{amsmath,amssymb,amsfonts}
\usepackage{subfigure}
\usepackage{algorithmic}
\usepackage{graphicx}
\usepackage{textcomp}
\usepackage{xcolor}
\usepackage{mathtools}
\usepackage{amsthm}
\usepackage{algorithm}
\usepackage{algorithmic}
\usepackage{hyperref}
\usepackage{booktabs, multirow} 
\usepackage{soul}
\usepackage{changepage,threeparttable} 

\def\BibTeX{{\rm B\kern-.05em{\sc i\kern-.025em b}\kern-.08em
    T\kern-.1667em\lower.7ex\hbox{E}\kern-.125emX}}
\begin{document}

\title{CR-VAE: Contrastive Regularization on Variational Autoencoders for Preventing Posterior Collapse}

\author{\IEEEauthorblockN{Fotios Lygerakis}
\IEEEauthorblockA{\textit{Chair of Cyber-Physical Systems} \\
\textit{University of Leoben}\\
Austria \\
fotios.lygerakis@unileoben.ac.at\\
https://orcid.org/0000-0001-8044-3511}
\and
\IEEEauthorblockN{Elmar Rueckert}
\IEEEauthorblockA{\textit{Chair of Cyber-Physical Systems}\\
\textit{University of Leoben}\\
Austria \\
elmar.rueckert@unileoben.ac.at\\
https://orcid.org/0000-0003-1221-8253}
}

\maketitle

\begin{abstract}
The Variational Autoencoder (VAE) is known to suffer from the phenomenon of \textit{posterior collapse}, where the latent representations generated by the model become independent of the inputs. 
This leads to degenerated representations of the input, which is attributed to the limitations of the VAE's objective function. 
In this work, we propose a novel solution to this issue, the \textit{Contrastive Regularization for Variational Autoencoders (CR-VAE)}. 
The core of our approach is to augment the original VAE with a contrastive objective that maximizes the mutual information between the representations of similar visual inputs. 
This strategy ensures that the information flow between the input and its latent representation is maximized, effectively avoiding \textit{posterior collapse}. 
We evaluate our method on a series of visual datasets and demonstrate, that CR-VAE outperforms state-of-the-art approaches in preventing \textit{posterior collapse}. Code for this project is available at \url{}https://github.com/ligerfotis/crvae.
\end{abstract}


\section{Introduction}
A compact, informative and robust representation of high dimensional data lies in the epicenter of the successful implementation of any algorithm in the realms of supervised~\cite{stacked_ae, CPC, moco} and reinforcement learning~\cite{curl, representations_rl, contrastive_rl}.
Such efficient representations encode the most relevant information, discarding or allocating small significance to the irrelevant one~\cite{stacked_ae, sparse_ae}.
In terms of robustness, the representation method must provide signal encodings that can be used independently of the underlying task and are invariant to the input variance~\cite{Li_2021_ICCV, AE_SDM,pmlr-v119-shi20e, VAE_LSPI, DBLP:journals/corr/abs-2007-13003}.

Thus, an efficient and robust representation can filter out the useful information that is hidden in the observed data and provide encodings that can generalize in a wider spectrum of applications.
Due to its power to both represent and generate data, the VAE is a popular method to obtain such representations. 
VAE is a latent variable graphical model that defines the joint distribution $p(x,z)$ of data $x$ and latent variable $z$. 
The latent variable $z$, which is distributed according to $p(z)$, greatly influences the generation of x via the likelihood $p(x|z)$. This is the generative part of the VAE. VAE has also a representation part, which consists of the posterior $p(z|x)$. Based on this analysis, VAE can factor $p(x,z)$ and obtain the approximated distribution of the data $p(x) = p(x|z)p(z)/p(z|x)$. This is achieved by approximating the likelihood $q_{\phi}(x|z)$ and 
\begin{figure}[h!]
    \centering
    \subfigure[Original ELBO]{\includegraphics[width=0.48\linewidth]{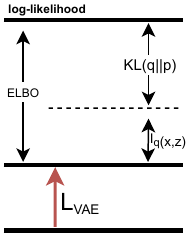}
    \label{fig:classic_elbo}}
    \subfigure[Contrastive Regularization]{\includegraphics[width=0.48\linewidth]{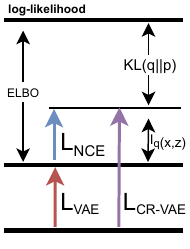}
    \label{elbo_augmented}}
    \caption{Visual description of the Evidence Lower BOund (ELBO) in a VAE. Figure~\ref{fig:classic_elbo} describes the minimization of ELBO, which consists of a mutual information and a Kullback–Leibler(KL) divergence term, as shown in Section~\ref{sec:post_colapse_origins}. The loss $\mathcal{L}_{VAE}$ (in red), minimizes ELBO, forcing $I_q$ to converge to zero, which leadis to \textit{posterior collapse}. The proposed CR-VAE loss $L_{CR-VAE}$ prevents that by augmenting $\mathcal{L}_{VAE}$ with a contrastive loss $\mathcal{L}_{NCE}$, putting a lower bound on $I_q(x;z)$ in Figure~\ref{elbo_augmented}.}
    \label{fig:graph_model}
\end{figure}
the posterior $q_{\theta}(z|x)$ with neural networks. The decoder network constitutes the approximated likelihood whereas the encoder constitutes the posterior.

%

The most critical drawback of the VAE is the \textit{posterior collapse}. When the decoder is expressive enough, something that happens when it is parameterized with a deep neural network, it becomes invariant to the latent representation. This phenomenon occurs because the decoder is powerful enough to approximate well the generative process of the training data, without considering the latent representation. Consequently, the generated data, even though it has low reconstruction error with the training samples, they are independent of the latent code. This leads the posterior to degrade and refrain it from encoding information from the input to the latent variable, causing what is called the \textit{posterior collapse}.

The contribution of this work is to tackle the \textit{posterior collapse} of VAE by augmenting its original objective with a new one. Specifically, we prove both theoretically and empirically that using a contrastive objective calculated between representations of similar and dissimilar inputs will reduce \textit{posterior collapse}. We call this method \textit{Contrastive Regularization for VAE (CR-VAE)}, the architecture of which is depicted in Figure~\ref{fig:grad_flow}.
We test our hypothesis on a series of datasets and measure the performance of the proposed method in a series of traditional measures, like negative log-likelihood, Kullback–Leibler(KL) divergence, mutual information, and active units. In addition, we measure the performance of the learned representations on common linear and distance-based instance discrimination tasks~\cite{learning_repr_max_mi, instance_level_discr,local_aggr, moco, simCLR, byol}. 
Finally, we visually examine the latent space of the learned representations, which we map on a two-dimensional space using t-SNE~\cite{tsne}.

\section{Related Work}
VAE is a popular unsupervised learning method based on neural networks that have been proposed by Kingma and Welling~\cite{vae} as a stochastic method to perform variational inference and learning.
To do so, they maximize the Evidence Lower BOund (ELBO) by simultaneously maximizing the log likelihood of the observed data $\log p_{\theta}(x)$ and minimizing the KL divergence between the approximate posterior $q_\phi(z|x)$ of the random variable $z$ and its prior $p(z
)$.
%

%
%
%
A major issue with the VAE is that if the decoder has the learning capacity to learn the exact data distribution $q(x)$, then the ELBO will be only dependent on $D_{KL}(q_\phi(z|x)||p(z))$. This will result to a \textit{posterior collapse}~\cite{post_collapse}, meaning that the encoder of the VAE will produce latent representations $z$ that will carry no information about input $x$. %

One approach to mitigate \textit{posterior collapse} has been focused on the architecture of the VAE itself or its training process. In \cite{saVAE} the authors combine the inference properties of amortized  variational inference with the good local properties of the stochastic one. Their semi-amortized uses the inference outputs of the first to initialize the latter. SkipVAE\cite{skipVAE} gives direct access to the input data for the generation of the latent variable using a skip connection between them. This allows the flow of information between them even in the case of \textit{posterior collapse}. Lagging VAE\cite{laggingVAE} spots the causes of \textit{posterior collapse} in the procedure of updating the parameters in a VAE. To solve what they call the “inference lag” they aggressively update the encoder before updating the whole model. That training scheme though comes with the pitfall of being computationally expensive.

Nevertheless, the series of studies that are most related to our work is the one that resorts to augmenting the ELBO objective with a mutual information penalty term.
Zhao et al tackle the \textit{posterior collapse} by introducing InfoVAE~\cite{infoVAE}. In their paper, they propose a new learning objective for the VAE, in which the preference between correct model inference and fitting the observed data distribution can be balanced by choosing a set of hyperparameters and dictating to the model the measure of reliance on the latent variables.
Another approach that tackles the posterior collapsing issue is the VAE-MINE~\cite{mutualVAE}. The authors introduce an additional learning objective for the VAE, which maximizes the mutual information between $x$ and $z$. VAE-MINE defines mutual information using the Jensen-Shannon divergence and computing the energy functions over the variational joint distribution.
The maximization of mutual information is also adopted in InfoMax-VAE~\cite{InfoMax} to improve the representation capabilities of a VAE. Here a dual form of the mutual information is adopted, using the convex conjugate of a broad class of divergences. InfoMax-VAE manages to outperform  InfoVAE and $\beta$-VAE in terms of encoding power comparing active units of the latent space.
A study that is the most related to ours was recently published by S Menon et al.~\cite{CriticVAE}. This work also capitalizes on contrastive learning to cure \textit{posterior collapse}. They are using a contrastive critic that detects the collapse by means of measuring the mutual information between the input and the latent representation and penalizes it. 
%

\section{Posterior Collapse in Variational Autoencoder}
This section introduces Variational Autoendoder (VAE) and explains the nature of \textit{posterior collapse}. We define the basic components of the variational inference model that constitutes the VAE, the notation of which can be found in Table~\ref{tab:naming}. Furthermore, posterior collapse is made visible through the exploration VAE's formulations of objective. By doing so, we can isolate the cause that leads to \textit{posterior collapse}, which integrated in VAE's objective.

\subsection{Variational Autoencoder (VAE)}
VAE~\cite{vae} is an amortized variational inference model that combines an inference and a generative model. It learns the parameters of the likelihood distribution $p(x|z)$, which constitutes the generative model, and approximates the posterior distribution $p(z|x)$ of the data $x$ over the latent variable $z$. The VAE uses a neural network with parameters $\phi$ and $\theta$ to approximate the joint distribution $q_{\theta, \phi}(x, z)$. The encoder approximates the posterior as $q_{\phi}(z|x)$ and the decoder the likelihood as $p_{\theta}(x|z)$. The goal of using a VAE as a generative model is to increase the probability of the data $p(x)$ by increasing the ELBO which is a known lower bound of the marginal likelihood. The prior $p(z)$ is typically chosen to be a multivariate normal Gaussian, and the encoder learns the parameters $\mu$ and $\sigma$ of the multivariate Normal distribution $p(z|x;{\mu}, {\sigma})$. Additionally, $z$ is selected to be a low-dimensional encoding of the input, forcing the encoder to learn underlying information in each input and discard irrelevant information.
\begin{table}\centering
\caption{Naming of referenced notation.}\label{tab:naming}
\begin{tabular}{ll}
\toprule
$p_{\theta, \phi}(x,z)$ & approximate joint distribution  \\
$p_{\theta}(x|z)$ & likelihood \\
$q_{\phi}(z|x)$ & approximate posterior \\
$p(z|x)$ & true posterior \\
$p(z)$ & prior \\
$q^{avg}_{\phi}(z)$ & average encoding distribution\\
$I_{q}(x, z)$ & mutual information of $x$ and $z$\\
$D_{KL}(q_\phi(z|x)||p(z))$ & KL divergence to prior\\
$D_{KL}(q^{avg}_\phi(z)||p(z))$ & marginal KL divergence to prior\\
\bottomrule
\end{tabular}
\end{table}
\subsection{The origins of \textit{posterior collapse}}
\label{sec:post_colapse_origins}
Posterior collapse is a phenomenon that occurs in VAEs where the latent variable $z$ does not carry any information about the input $x$. Its origins can be found in maximiazing the ELBO objective, which is used for training VAEs. 
ELBO consists of two terms: the reconstruction error denoted by $logq_{\theta}(x|z)$ and the KL divergence between the posterior and the prior. The two terms form the ELBO as: 
\begin{equation}
\begin{split}
    & ELBO(\theta, \phi) 
    = \mathbb{E}_{q(x)}[ \mathbb{E}_{q_{\phi}(z|x)}[\log p_{\theta}(x|z)]] \\
    & - D_{KL}(q_\phi(z|x)||p(z)) \;,
\end{split}
 \label{elbo_equation}
\end{equation}
where $q(x)$ is the empirical data distribution, drawn from the dataset the model is trained on. The KL divergence term aims to make the posterior similar to the prior, which can lead to $q_{\phi}(z|x)$ becoming equal to $p(z)$ and the latent variable $z$ not representing the input $x$ at all, resulting in \textit{posterior collapse}~\cite{lossy_VAE, e3468bd4487f44458c33dd7ed459fdb1, NIPS2016_ddeebdee}. 

%
Another way of identifying \textit{posterior collapse} is to monitor the mutual information $I_{q}(x, z)$ between $x$ and $z$. \textit{Posterior collapse} is apparent when $I_{q}(x, z)$ is zero. In this case $z$ is not encoding the input $x$ at all. 
It can be proven that there exists a lower bound for the mutual information between the input x and its representation $z$.
\begin{equation} \label{eq1}
\begin{split}
	& \mathbb{E}_{q(x)} [D_{KL}(q_\phi(z|x)||p(z))], \\
    & = \int_{x \sim q(x)}^{} \int_{z \sim q_\phi(z|x)}^{} q_\phi(x,z) \; \log\frac{q_\phi(z|x)}{p(z)} dz dx\;. \nonumber
\end{split}
\end{equation}
As KL divergence is always non-negative, we can define the following inequality:
\begin{equation}
\begin{split}
    &\mathbb{E}_{q(x)} [D_{KL}(q_\phi(z|x)||p(z))] \\
    &\geq \int_{x,z}^{} q_\phi(x,z) log\frac{q_\phi(z|x)}{p(z)} dz dx - D_{KL}(q_\phi(z)||p(z)), \\
    & = \int_{x,z}^{} q_\phi(x,z) \left[log\frac{q_\phi(z|x)}{p(z)} - log\frac{q_\phi(z)}{p(z)}\right] dz dx, \\
    & = \int_{x,z}^{} q_\phi(x,z) log\frac{q_\phi(z|x)}{q_\phi(z)} dz dx, \\
    & = \int_{x,z}^{} q_\phi(x,z) log\frac{q_\phi(x,z)}{q_\phi(x)q_\phi(z)} dz dx = I_q(x;z) \; \nonumber
\end{split}
\end{equation}
This inequality proves that minimizing the second part of the ELBO objective (Equation~\ref{elbo_equation}) leads to minimizing the information that $z$ shares with $x$. In \cite{InfoMax} authors use this inequality to tackle \textit{posterior collapse} by calculating a dual form of $I_{q}(x, z)$ and adding it as a weighted loss to the original training objective.

In addition, in ~\cite{elbo_surgery} Equation~\ref{elbo_equation} is further dissected into:
\begin{equation}
\begin{split}
    & ELBO(\theta, \phi) 
    = \mathbb{E}_{q(x)}[ \mathbb{E}_{q_{\phi}(z|x)}[logq_{\theta}(x|z)]] \\
    & - I_{q}(x, z) - D_{KL}(q^{avg}_\phi(z)||p(z)) \;,
\end{split}
 \label{elbo_dissection}
\end{equation}
where the KL divergence term has been factored into a mutual information term between $x$ and the latent variable $z$, and a marginal KL to prior term between the average encoding distribution and the true prior of $p(z)$. From that, we see an alternative approach to the original objective of the VAE, which leads to the reduction of $I_{q}(x, z)$.

To mitigate this effect, we propose the augmentation of VAE's objective with an additional term that maximizes the aforementioned mutual information. To do that, we need to augment the original VAE loss with a term that acts adversarially in the minimization of $I_{q}(x, z)$. We introduce such an objective in the following section.

From this point, we use the notation KL to refer to $D_{KL}(q_\phi(z|x)||p(z))$ to declutter notation. Any other KL divergence term will be referenced descriptively. We also use the mathematical term $D_{KL}$ to descibe the KL divergence.

\section{Contrastive Regularized Variational Autoencoder (CR-VAE)}
We solve the problem of \textit{posterior collapse} by finding and maximizing a known lower bound of $I_{q}(x, z)$. van den Oord et al.(\cite{CPC}) have shown that the InfoNCE loss $\mathcal{L}_{InfoNCE}$, a widely used loss in contrastive representation learning since then, constitutes a lower bound for $I_{q}(x, z)$:
\begin{equation} \label{nce_MI}
    I_{q}(x, z) \geq \log(K) - \mathcal{L}_{InfoNCE} \;,
\end{equation}
where $K$ is the batch size. Our goal thus becomes to augment VAE's original loss with $\mathcal{L}_{InfoNCE}$, which will act as a regularizer to $I_{q}(x, z)$ term of Equation~\ref{elbo_dissection} and preventing \textit{posterior collapse}.

The InfoNCE loss, which is based on the Noise Contrastrive Estimation (NCE) method~\cite{nce}, utilizes the categorical cross-entropy loss function to distinguish the positive sample, from a group of unrelated negative samples, found in the batch. The InfoNCE loss is computed as:
\begin{equation} \label{infonce}
\mathcal{L}_{InfoNCE} = - \mathbf{E}\left[log \cfrac{e^{sim(z_i, z_j)}}{{\sum_{k=1}^{K} e^{sim(z_{i}, z_{k})}}}\right] \;,
\end{equation}
where $sim(\cdot)$ is the cosine distance similarity measure, $z_i$ and $z_j$ are the representations of a positive pair and $z_k$ the representations of the negative ones. 

\subsection{Fixing ELBO with Contrastive Regularization}
We construct the loss function of our CR-VAE by combining the original ELBO objective in Equation~\ref{elbo_equation} with the contrastive loss $\mathcal{L}_{InfoNCE}$~\cite{nce}, weighted with a hyperparameter $\gamma$, into:
\begin{equation}
	\mathcal{L}_{CR} = \mathcal{L}_{VAE} + \gamma \cdot \mathcal{L}_{InfoNCE}\;, \nonumber
\end{equation}
where $\mathcal{L}_{VAE} = - ELBO$.
We learn the parameters $\phi$ and $\theta$ of $q_{\phi}$ and $p_{\theta}$ such that we minimize $\mathcal{L}_{CR-VAE}$:
\begin{equation}
	\theta*, \phi* = \arg \min_{\theta, \phi} \mathcal{L}_{CR-VAE} \; .  \nonumber
\end{equation} 
These parameters are learned through the gradients of the $\mathcal{L}_{CR-VAE}$ loss using backpropagation. (See Section~\ref{sec:model_params})

In a recent work ~\cite{CriticVAE}, \textit{posterior collapse} is tackled by augmenting the original ELBO objective with a loss that maximizes the lower bound of $I_q(x;z)$ too, obtained by a contrastive critic. In our work, we compute the InfoNCE loss using the encoder of the VAE to compute the lower bound of $I_{q}(x, z)$. In contrast,~\cite{CriticVAE} train a separate softmax classifier to detect \textit{posterior collapse} and penalize against it by augmenting the original loss the estimated InfoNCE loss. As we show in Section~\ref{sec:results}, CR-VAE yields better results in tackling \textit{posterior collapse}. In addition, CR-VAE constitutes a simpler implementation with minimal overhead over the original VAE, as we discuss in the folowing section.

\subsection{CR-VAE Model Architecture}
One of the advantages of our method is the simplicity in its implementation. In this paper, we consider the approach of the Momentum Contrast (MoCo) by~\cite{moco} to compute $\mathcal{L}_{InfoNCE}$. Therefore, the original VAE's encoder is used to compute the query representations $z_q$ and a duplicate encoder is used to encode the key representations $z_k$, as shown in Figure~\ref{fig:grad_flow}. The parameters of the key encoder are computed using the exponential moving average of the query encoder's weights. We obtain queries and keys using random data augmentation of the input $g(x)$, as  described in the Technical Appendix.

\section{Experiments}
In this section, we evaluate CR-VAE and compare it to the traditional VAE (VAE) and a series of baseline approaches. We first set the scene of the experimental setup and then we discuss the metrics we use to detect and measure \textit{posterior collapse}. 

\subsection{Experimental Setup}
\label{sec:experimental_setup}
We evaluate CR-VAE on the MNIST~\cite{mnist}, EMNIST~\cite{emnist}, FashionMNIST~\cite{fashionmnist}, CIFAR10\cite{cifar} and Omniglot~\cite{omniglot} datasets on a standard experimental setup across all of them. All baselines in Table~\ref{tab:results2} are implemented using the same network architecture and training hyperparameters. This is a common benefit of using ELBO-augmentation methods for tackling \textit{posterior collapse} ~\cite{InfoMax, infoVAE, CriticVAE}. To compare our method with other baselines, we either augment the ELBO of our VAE implementation with the respective loss of the method (Table~\ref{tab:results2}) or build upon other baseline implementations with our CR-VAE objective (Table~\ref{tab:omniglot_comparison}). The choice of the objective weights for each baseline was made according to the respective publication instructions. 
Furthermore, we use the Mean Squared Error as reconstruction loss in our VAE implementation. We backpropagate the error using stochastic gradient descent with a learning rate of $0.001$. More details for the reconstruction loss and training hyper-parameters can be found in the Technical Appendix.

Experiments on MNIST, EMNIST, FashionMNIST and CIFAR10 use a 16-dimensional latent space and a batch size of 256. High batch size is essential for CR-VAE. From Equation~\ref{nce_MI} we see that the value of mutual information bound depends on $\log K$. Using $K=256$, we can reach a theoretical lower bound of $\log_{2}(256)=8$ bit for mutual information, while using a batch size of $50$, the theoretical lower bound is $\log_{2}(50)\approx5.64$ bits. For the comparison in Table~\ref{tab:omniglot_comparison} we use the same hyperparameters as the baselines. There, the batch size is selected to be 50, leading to smaller bounds for the mutual information, as confirmed by the experimental results.
\begin{figure}
\centering
  \includegraphics[width=1\linewidth]{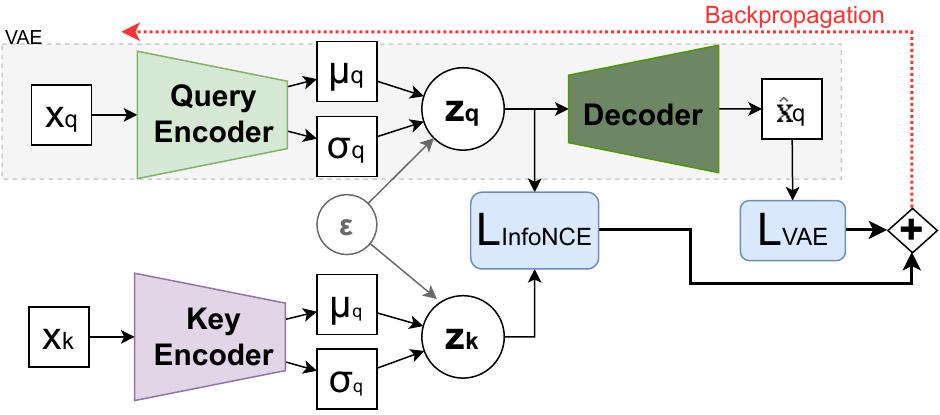}  
    \caption{Illustration of the proposed Contrastive Regularised Variational Autoencoder (CR-VAE) architecture that builds on the original VAE (shaded in grey). 
    The proposed contrastive objective $\mathcal{L}_{InfoNCE}$ prevents the \textit{posterior collapse} of the query encoder. The random variable $\epsilon$ is normally distributed and is used for the reparameterization trick~\cite{vae}.}
    \label{fig:grad_flow}
\end{figure}
\subsection{Model Architecture.}
\label{sec:model_architecture}
For the experiments presented in Table~\ref{tab:results2}, we utilized a custom architecture with an encoder that has one less hidden layer than the decoder. This resulted in an encoder with $40\%$ fewer parameters. By doing so, we aimed to mimic the \textit{posterior collapse} that occurs in powerful generative models. Additionally, we avoided tackling \textit{posterior collapse} due to architectural choices by not using skip connections between the layers, as recommended in ~\cite{skipVAE}.

For the encoder, we utilized a small custom architecture that consisted of 3 convolutional layers and a fully connected layer that projects from the feature maps to the latent space dimensions, totaling approximately 176K parameters. On the other hand, the decoder consisted of 4 convolutional layers and a total of 439K parameters, achieving higher capacity and bigger expressive power than the encoder.

It is important to mention that if the encoder is constructed as an architecturally mirrored version of the decoder, then the original VAE can learn multiple distinct modes in the latent space for the MNIST and EMNIST datasets. However, this is not the case for every architecture, hyperparameter, or optimizer. In our study, we intentionally chose the aforementioned architecture and used SGD instead of Adam as an optimizer to demonstrate more clearly the contribution of our proposed CR-VAE method in addressing \textit{posterior collapse}.

To perform a fair comparison with the rest of the baselines on Table~\ref{tab:omniglot_comparison}, we adopt the architecture and training hyper-parameter choices described in the literature~\cite{CriticVAE, laggingVAE, skipVAE, saVAE}. 

\subsection{CR-VAE Model Learning.}
\label{sec:model_params}
We compute the gradients of the encoder parameters $\phi$ and the decoder parameters $\theta$ w.r.t. $\mathcal{L}_{ELBO}$ and  $\mathcal{L}_{InfoNCE}$, i.e. $\nabla \mathcal{L}_{ELBO} (\phi)$ $\nabla \mathcal{L}_{InfoNCE} (\phi)$ and $\nabla \mathcal{L}_{ELBO} (\theta)$ $\nabla \mathcal{L}_{InfoNCE} (\theta)$ respectively. Then we backpropagate the combination of the gradients w.r.t $\theta$ through both the decoder and the encoder parameters (See Figure~\ref{fig:grad_flow})
The parameters of the key encoder $k_{\phi}$ are updated using the Exponential Moving Average(EMA) of the query encoder $q_{\phi}$. 
Algorithm~\ref{alg:CR-VAE}, presents the core components and update rules of CR-VAE. 

\begin{algorithm}
\caption{CR-VAE}\label{alg:CR-VAE}
\begin{algorithmic}

\STATE Initialize parameters $\phi_{q}$ of query encoder $p_{q}(z|x;\phi)$
\STATE Initialize parameters $\theta_{q}$ of query decoder $p_{q}(x|z;\theta)$
\STATE Initialize parameters $\phi_{k} \gets \phi_{q}$ of query encoder $f_{k}(x;\phi)$
\FOR{$e$ in $epochs$} 
    \FOR{$x$ in $train$ $loader$} 
        \STATE $x_{q} \gets g(x)$
        \STATE $x_{k} \gets g(x)$
    
        \STATE $z_{q} \sim p_{q}(.|x;\phi)$
        \STATE $z_{k} \sim p_{k}(.|x_{k};\phi)$
        
        \STATE $\mathcal{L}_{InfoNCE} \gets InfoNCE(z_{q}, z_{k})$
        
        \STATE $\hat{x_{q}} \gets p_{q}(.|z_{q};\theta)$
        \STATE $\mathcal{L}_{ELBO} \gets \log p(\hat{x}) - D_{KL}(p_{q}(z_{q}|x;\phi)||p(z_{q}))$
        
        \STATE $\phi_{q} \gets \phi_{q} + lr * (\nabla \mathcal{L}_{ELBO} (\phi) + \gamma * \nabla \mathcal{L}_{InfoNCE} (\phi))$
        \STATE $\theta_{q} \gets \theta_{q} + lr * (\nabla \mathcal{L}_{ELBO} (\theta) + \gamma * \nabla \mathcal{L}_{InfoNCE} (\theta))$
    \ENDFOR
    \STATE $\phi_{k} \gets EMA(\phi_{q})$
\ENDFOR
\end{algorithmic}
\end{algorithm}

\subsection{Performance Metrics}
Unfortunately, there is not yet in the literature a direct measure of \textit{posterior collapse}. Therefore the related work has to infer it or measure it with indirect measures. Hence, to measure the impact of CR-VAE in tackling \textit{posterior collapse}, we evaluate the learned representations with four metrics commonly used in the related work. 
\begin{table*}[!htp]\centering
\caption{Comparison of CR-VAE with previous state-of-the-art approaches for solving \textit{posterior collapse} on Omniglot dataset. The results of the baseline approaches were referenced from~\cite{saVAE},~\cite{skipVAE},~\cite{laggingVAE} and~\cite{CriticVAE}. The computation of NLL for CR-VAE was held out with the use of MSE. Mean and standard deviation in parenthesis over $5$ independent runs.}
\label{tab:omniglot_comparison}
\scriptsize
\begin{tabular}{lccccc}\toprule
Model &NLL &KL &MI &AU \\\midrule
VAE &89.41(0.04) &1.51(0.05) &1.43(0.07) &3(0.0) \\
SA-VAE\cite{saVAE} &89.29 (0.04) &2.55 (0.05) &2.20 (0.03) &4.0 (0.0) \\
Skip-VAE\cite{skipVAE} &89.41 (0.05) &1.75 (0.20) &1.61 (0.10) &3.0 (0.4) \\
Lagging-VAE\cite{laggingVAE} &89.05 (0.05) &2.51 (0.14) &2.19 (0.08) &5.6 (0.5) \\
VAE + Contrastive Critic~\cite{CriticVAE} &89.24 (0.05) &7.66 (0.14) &3.82 (0.04) &\textbf{28.0} (0.0) \\
CR-VAE &\textbf{25.96}(0.83) &\textbf{5.64}(1.66) &\textbf{4.62}(0.69) &{26.67}(1.55) \\
CR-VAE (batch size $256$) &\textbf{25.04}(0.47) &\textbf{6.78}(1.49) &\textbf{5.00}(0.07) &{26.00}(1.73) \\
\bottomrule
\end{tabular}
\end{table*} 
\begin{table*}[!htp]\centering
\caption{Results on the MNIST, EMNIST, FashionMNIST and CIFAR10 datasets. Mean value over $5$ independent runs. Standard deviation in parenthesis.}\label{tab:results2}
\scriptsize
\begin{tabular}{lrcccccccc}\toprule
Dataset &Model &NLL &KL &$KL(q(z)||p(z))$ &MI &AU &Linear &KNN \\\midrule
\multirow{4}{*}{MNIST} &VAE &89.30(0.01) &13.16(0.09) &10.75(0.06) &2.41(0.03) &0.00(0.00) &12.29(1.52) &33.12(1.75) \\
&InfoVAE &89.30(0.01) &13.18(0.07) &10.75(0.03) &2.43(0.04) &0.00(0.00) &13.11(1.80) &32.98(1.23) \\
&InfoMax-VAE &75.45(0.31) &12.75(0.10) &\textbf{10.38(0.07)} &2.37(0.03) &0.00(0.00) &82.77(2.04) &28.66(1.60) \\
&CR-VAE &\textbf{22.24(0.10)} &\textbf{16.99(0.08)} &12.30(0.04) &\textbf{4.69(0.04)} &\textbf{16.00(0.00)} &\textbf{88.51(0.92)} &\textbf{95.29(0.14)} \\\\\midrule
\multirow{4}{*}{EMNIST} &VAE &111.35(0.01) &13.02(0.08) &\textbf{10.07(-0.73)} &2.95(0.81) &5.33(8.26) &2.74(0.03) &14.73(0.86) \\
&InfoVAE &111.35(0.01) &13.20(0.05) &10.76(0.04) &2.44(0.01) &5.12(7.81) &2.83(0.18) &14.73(0.76) \\
&InfoMax-VAE &120.00(0.34) &12.96(0.10) &10.55(-0.60) &2.41(0.70) &3.20(7.15) &11.22(1.84) &25.31(0.17) \\
&CR-VAE &\textbf{22.44(0.27)} &\textbf{18.08(0.43)} &13.88(0.21) &\textbf{4.20(0.22)} &\textbf{16.00(0.00)} &\textbf{52.90(2.40)} &\textbf{77.83(0.29)} \\\midrule
\multirow{4}{*}{FashionMNIST} &VAE &\textbf{16.02(0.25)} &8.41(0.09) &5.28(0.02) &3.13(0.07) &12.33(1.97) &62.03(3.67) &80.74(0.33) \\
&InfoVAE &17.23(0.16) &8.33(0.09) &\textbf{4.92(-0.03)} &3.41(0.12) &12.48(1.11) &63.02(1.02) &79.42(0.55) \\
&InfoMax-VAE &16.04(0.33) &8.41(0.10) &5.26(0.03) &3.15(0.07) &12.80(1.94) &61.87(1.57) &81.28(0.28) \\
&CR-VAE &21.73(0.26) &\textbf{10.40(0.07)} &6.82(-0.15) &\textbf{3.58(0.22)} &\textbf{16.00(0.00)} &\textbf{68.37(3.21)} &\textbf{82.29(0.72)} \\\midrule
\multirow{4}{*}{CIFAR10} &VAE &60.90(0.24) &18.36(0.13) &14.92(0.01) &3.44(0.12) &16.00(0.00) &18.75(2.57) &42.32(0.12) \\
&InfoVAE &\textbf{58.49(0.24)} &18.45(0.31) &14.42(0.22) &3.53(0.09) &16.00(0.00) &18.06(1.71) &42.12(0.62) \\
&InfoMax-VAE &60.19(1.06) &19.43(4.30) &16.04(4.00) &3.39(0.30) &16.00(0.00) &16.38(4.41) &35.72(0.55) \\
&CR-VAE &62.34(0.63) &\textbf{16.76(0.63)} &\textbf{13.22(0.51)} &\textbf{3.54(0.12)} &\textbf{16.00(0.00)} &\textbf{20.78(0.72)} &\textbf{43.60(0.51)} \\
\bottomrule
\end{tabular}
\end{table*}

\subsubsection{Negative Log Likelihood(NLL).}
\label{sec:nll}
Negative Log Likelihood (NLL) consists of the reconstruction capability of a VAE. This is a very intuitive metric for any kind of VAE since it consists of one of the two basic metrics upon which one is optimized. In this work, we use the Mean Squared Error (MSE) between the input and its reconstruction from the decoder to measure the reconstruction error. We specifically choose MSE as we define the likelihood $p(x|z)$ to be an isotropic Gaussian distribution $\mathcal{N}(\mu, \sigma^2\mathcal{I})$. We shall note that not all baselines have adopted this design choice. 
All the methods on Table~\ref{tab:omniglot_comparison}, including the original VAE on this table, have assumed a Bernoulli-distributed observations. For this reason, they use the Binary Cross Entropy (BCE) loss. (For details on the derivations of the reconstruction losses see the Technical Appendix.)
\subsubsection{KL Divergence to prior (KL).}
Keeping track of KL is important for observing \textit{posterior collapse}. When the posterior has collapsed, the ELBO is kept relatively high, but KL converges to zero. Thus, the loss is dominated by the $D_{KL}(q_\phi(z|x)||p(z))$ term and the NLL error stays high.
\subsubsection{Mutual Information (\textit{MI}).} 
Similarly to ~\cite{elbo_surgery} we indirectly calculate the mutual information (MI) term of Equation~\ref{elbo_dissection}. First, we calculate the $D_{KL}(q_\phi(z)||p(z))$ term via ancestral sampling and then we subtract it from the original KL term of the ELBO to obtain MI. The more the MI is, the more information is shared between the observed input and its representation, indicating the mitigation of \textit{posterior collapse}.
\subsubsection{Active Units.} 
The Active Units (AU) \cite{activation_units} metric measures the subset of latent variables used actively to generate the reconstructed observation by the decoder. More specifically, AU measures the result of variance of the input on the latent variables. A collapsed posterior would mean that the latent units do not respond to the changes of the input. The AU value is given by $\text{AU}(z) = \text{Cov}_{x}(\mathbb{E}_{z \sim q(z|x)}[z])$, and we consider that a unit is activated if is has an AU value greater than a threshold of $0.01$.

\subsubsection{Latent Space Clustering.} 
On top of the traditional metrics above, we propose the visualization of the latent space using t-distributed stochastic neighbor embedding (t-SNE)~\cite{tsne} to infer whether \textit{posterior collapse} occurs. By visualizing the latent space using t-SNE, one can observe if the encoded data forms distinct clusters. This can indicate that the VAE is encoding meaningful information, or if the encoded data is spread out or overlapped, that \textit{posterior collapse} may be occurring.
\subsubsection{Linear Classification (\textit{Linear}).} 
In the absence of \textit{posterior collapse}, the latent representations will carry important information. The more the amount of information is carried throughout the input in the compact representation, the better the results they will yield in a  discrimination task. The most straightforward and common way to evaluate the representation capabilities of a method among the representation learning community ~\cite{moco, simCLR, data-effecient-cpc, pretext_invariant_repr} has been the linear classification protocol. 
Thus, in this work, we use the learned representation of a VAE to measure how well it tackles \textit{posterior collapse}. 
Our hypothesis here is that if the representations encode enough information about the input, they will lead the simple linear classifier (\textit{Linear}) to improved classification accuracy in comparison to methods where \textit{posterior collapse} is not tackled properly.
We thus freeze the encoder weights and train a fully-connected layer followed by a softmax activation. We train the classifier on the frozen representations for $200$ epochs with a learning rate of $10^{-3}$ and report the top-1 classification accuracy~\cite{moco}.

\subsubsection{Distance-based Classification (\textit{KNN}).} 
In addition to the \textit{Linear} classification, we examine the performance of the learned features on a distance-based classifier, having the same hypothesis with the previous paragraph in mind. 
For this task, we use the K-Nearest Neighbor (\textit{KNN}) algorithm that classifies a data instance based on the majority vote of the K closest neighbors. 
The metric to measure proximity to a neighbor is the distance-weighted cosine distance. 
With this measure, we seek to evaluate both the efficiency of the learned features and their robustness. \textit{KNN} classification performance informs us whether representations of similar inputs are close in the latent space. Furthermore, it provides a measure of the robustness of the variance of the input.

\subsection{Results}
\label{sec:results}
We compare CR-VAE against other state-of-the-art approaches in terms of the common literature metrics as well as the two semi-supervised tasks. Finally, we examine the structure of the learned latent space.
We report the experimental results in Tables~\ref{tab:results2} and~\ref{tab:omniglot_comparison}, where the effectiveness of CR-VAE in tackling \textit{posterior collapse} is clearly indicated. 

\subsubsection{Comparing with the Baseline}
On Table~\ref{tab:results2} we report the results on held out data of the original VAE, the InfoVAE~\cite{infoVAE}, the InfoMax-VAE~\cite{InfoMax} and CR-VAE. Our method consistently achieves the highest MI between the input and its representation across all datasets. It is worth noticing, that even in the absence of \textit{posterior collapse} by the original VAE on the FashionMnist and CIFAR10 datasets, CR-VAE was able to keep higher MI. In addition, CR-VAE consistently achieves the highest AU, Linear and KNN classification scores.

An important notice about CR-VAE is the choice of the batch size while training, as we discussed on Section~\ref{sec:experimental_setup}. As we see on Table~\ref{tab:omniglot_comparison}, even though CR-VAE is outperforming the baseline with the common batch size of $50$ samples, it provides an even better ELBO with a higher MI with when trained with a batch size of $256$ samples.
\subsubsection{Semisupervised tasks}
Nevertheless, the representation provided by CR-VAE seems to be better regularized and allows more information to flow from the input to the representation. 
This can be concluded from the both the \textit{Linear} and \textit{KNN} classification performance results across the MNIST, EMNIST, FashionMNIST and CIFAR10 datasets. We observe an intuitive correlation between a methods maintaining high MI and having high accuracy on the downstream classification tasks. CR-VAE representations lead to higher classification performance for the datasets, where \textit{posterior collapse} is present in the VAE.  
This is another support to the conclusion that CR-VAE avoids the \textit{posterior collapse} more efficiently than other approaches and maintains enough information and structure in the latent encoding that leads to high classification.

\subsubsection{Ablation study on CR-VAE}
\begin{figure}
  \includegraphics[width=1\linewidth]{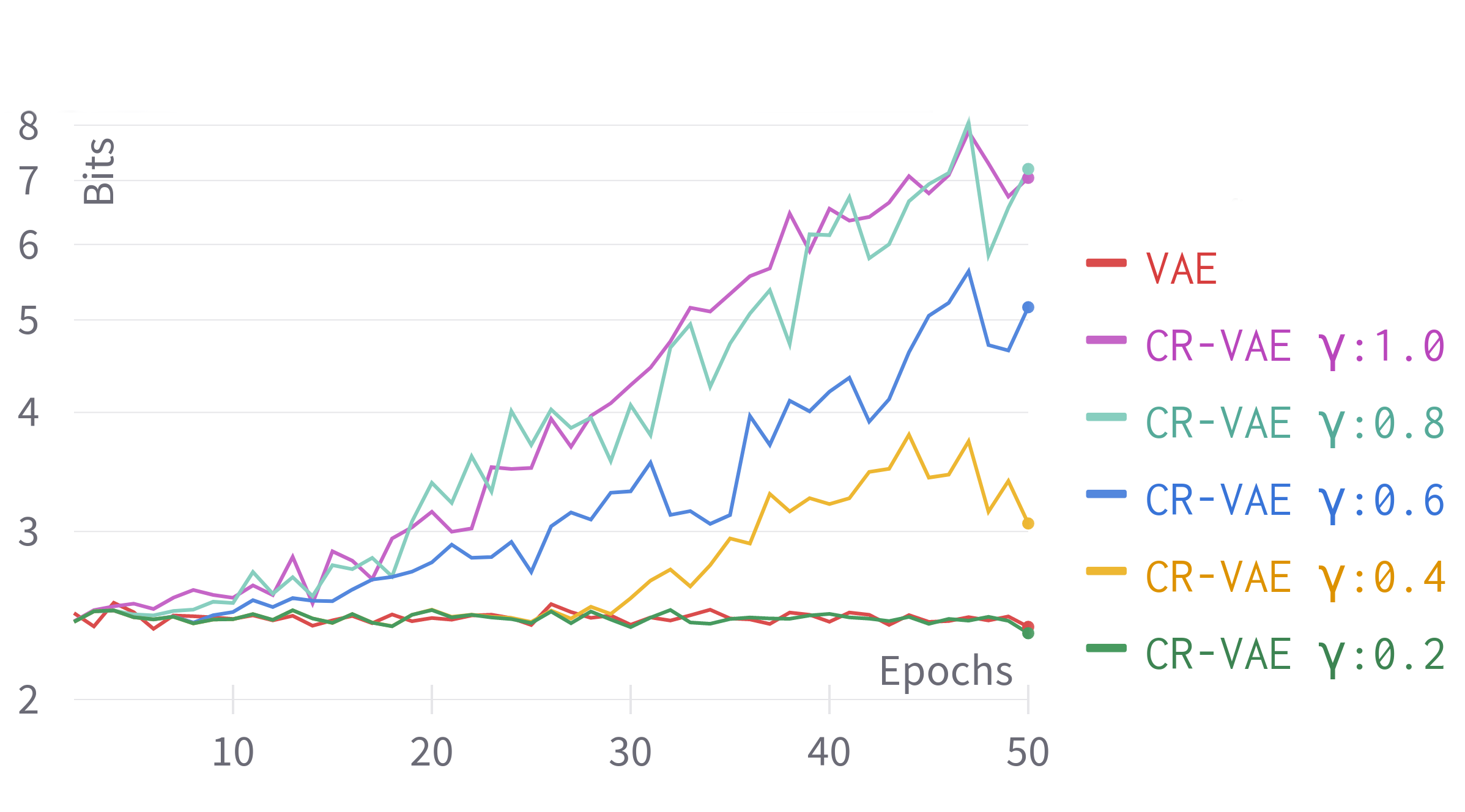}  
    \caption{The MI measured on held-out data during the training of CR-VAE and original VAE. We ablate the $\gamma$ values for the contrastive regularization for VAE. Values are shown in a logarithmic scale. The curves for VAE and CR-VAE with $\gamma=0.0$ and $\gamma=0.2$ overlap and thus not easily distinguishable.}
    \label{fig:ablation}
\end{figure}

We perform an ablation study on different values of $\gamma$, to show the effect of contrastive regularization for increasing the lower bound on the MI of the learned model. 
\begin{figure}
    \centering
    \subfigure[Original VAE latent of MNIST]{\includegraphics[trim={0 0 6.5cm 2.5cm},clip,width=0.48\linewidth]{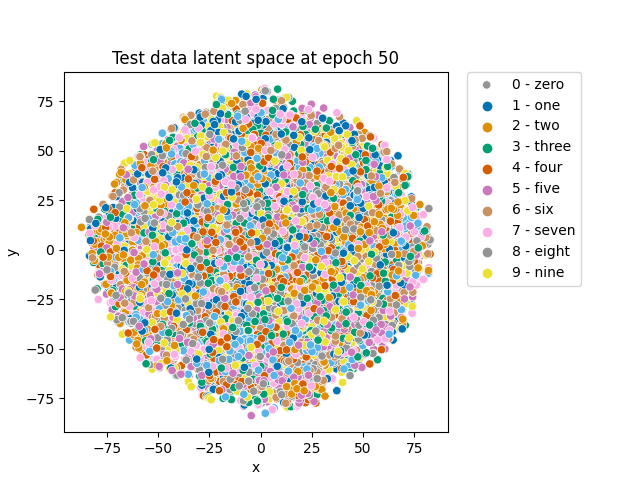}    
    \label{fig:mnist_latent_vae}}
    \subfigure[CR-VAE latent of MNIST]{\includegraphics[trim={0 0 6.5cm 2.5cm},clip,width=0.48\linewidth]{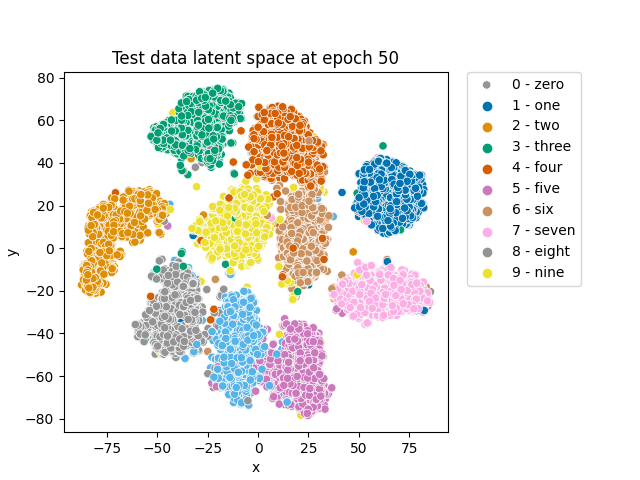}
    \label{fig:mnist_latent_crvae}}
    \subfigure[Original VAE latent of EMNIST]{\includegraphics[trim={0 0 6.5cm 2.5cm},clip,width=0.48\linewidth]{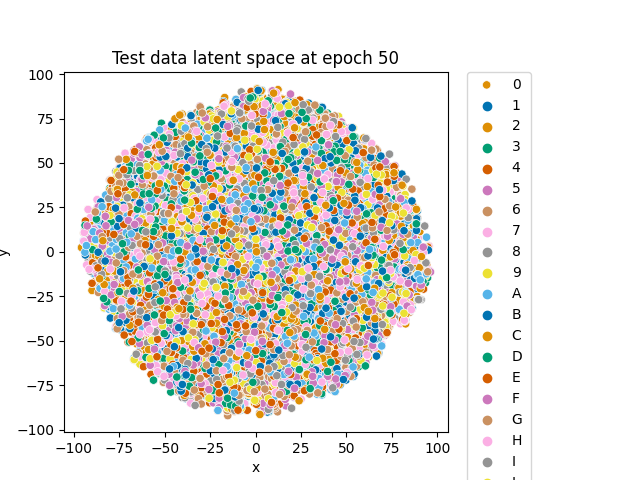}
    \label{fig:emnist_latent_vae}}
    \subfigure[CR-VAE latent of EMNIST]{\includegraphics[trim={0 0 6.5cm 2.5cm},clip,width=0.48\linewidth]{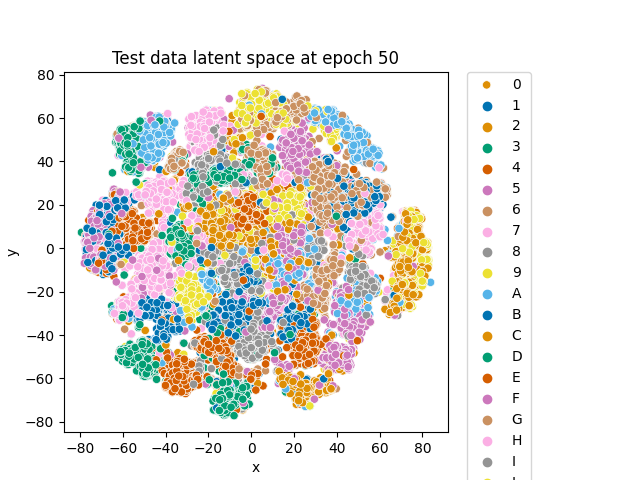}
    \label{fig:emnist_latent_crvae}}
    \subfigure[VAE latent of FashionMNIST]{\includegraphics[trim={0 0 6.5cm 2.5cm},clip,width=0.48\linewidth]{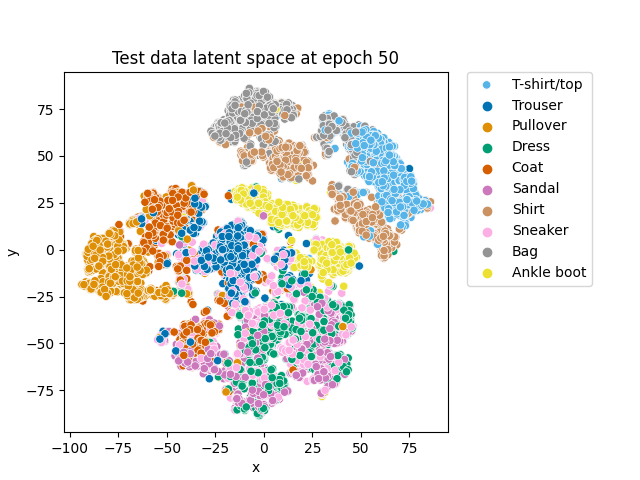}
    \label{fig:fashionmnist_latent_vae}}
    \subfigure[CR-VAE latent of FashionMNIST]{\includegraphics[trim={0 0 6.5cm 2.5cm},clip,width=0.48\linewidth]{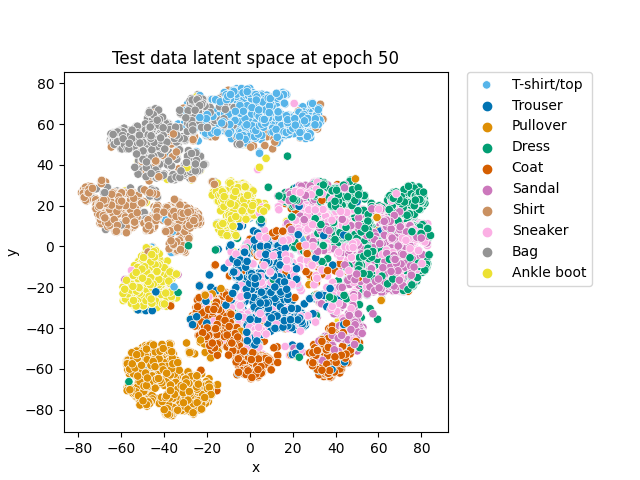}
    \label{fig:fashionmnist_latent_crvae}}
    \subfigure[VAE latent of CIFAR10]{\includegraphics[trim={0 0 6.5cm 2.5cm},clip,width=0.48\linewidth]{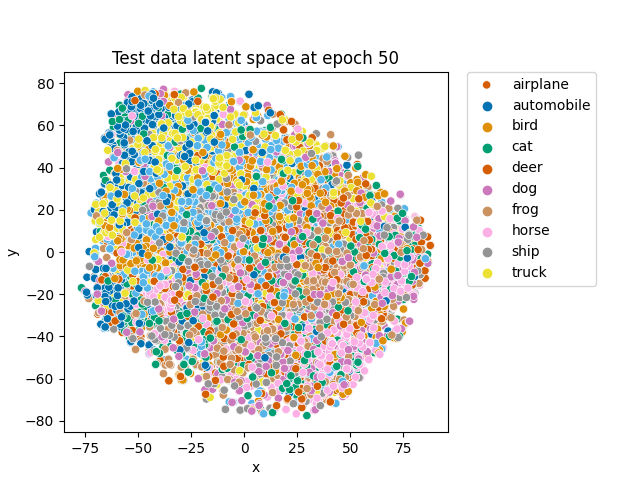}
    \label{fig:cifar_latent_vae}}
    \subfigure[CR-VAE latent of CIFAR10]{\includegraphics[trim={0 0 6.5cm 2.5cm},clip,width=0.48\linewidth]{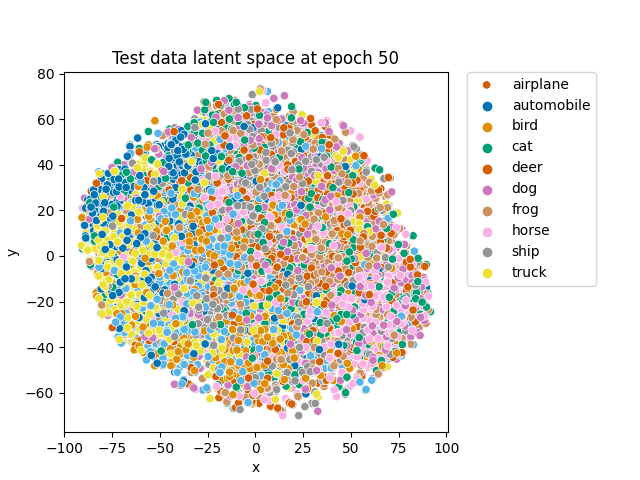}
    \label{fig:cifar_latent_crvae}}
    \caption{The learned latent space of the MNIST, EMNIST, FashionMNIST and CIFAR10 datasets mapped on a 2d space using t-SNE~\cite{tsne}. Posterior collapse is evident for VAE (left column) on MNIST and EMNIST datasets, while CR-VAE (right column) shows distinct embedding clusters. Due to the flat architecture and hyperparameter choice across all datasets, no collapse is observed for the FashionMNIST dataset, whereas for the CIFAR-10 complete collapse is not observed. Yet, the latent space of both VAE and CR-VAE for CIFAR-10, under this experimental set-up, present modes that are mixed and not completely separable from each other.}
        \label{fig:latent_2d}
\end{figure}
We also show that it is possible to retrieve the original VAE from CR-VAE. We start with $\gamma=0$ and increase it with a step size of $0.2$ until $\gamma=1$ to train CR-VAE. From the curves in Figure~\ref{fig:ablation}, we see that for $\gamma=0$ and $\gamma=0.2$, CR-VAE has the same MI as VAE. For bigger values of $\gamma$, the MI of CR-VAE keeps increasing analogously to the value of $\gamma$.
\subsubsection{Latent Space Structure}
\begin{figure}[!h]
    \centering
    \subfigure[Original VAE on MNIST]{\includegraphics[width=0.48\linewidth]{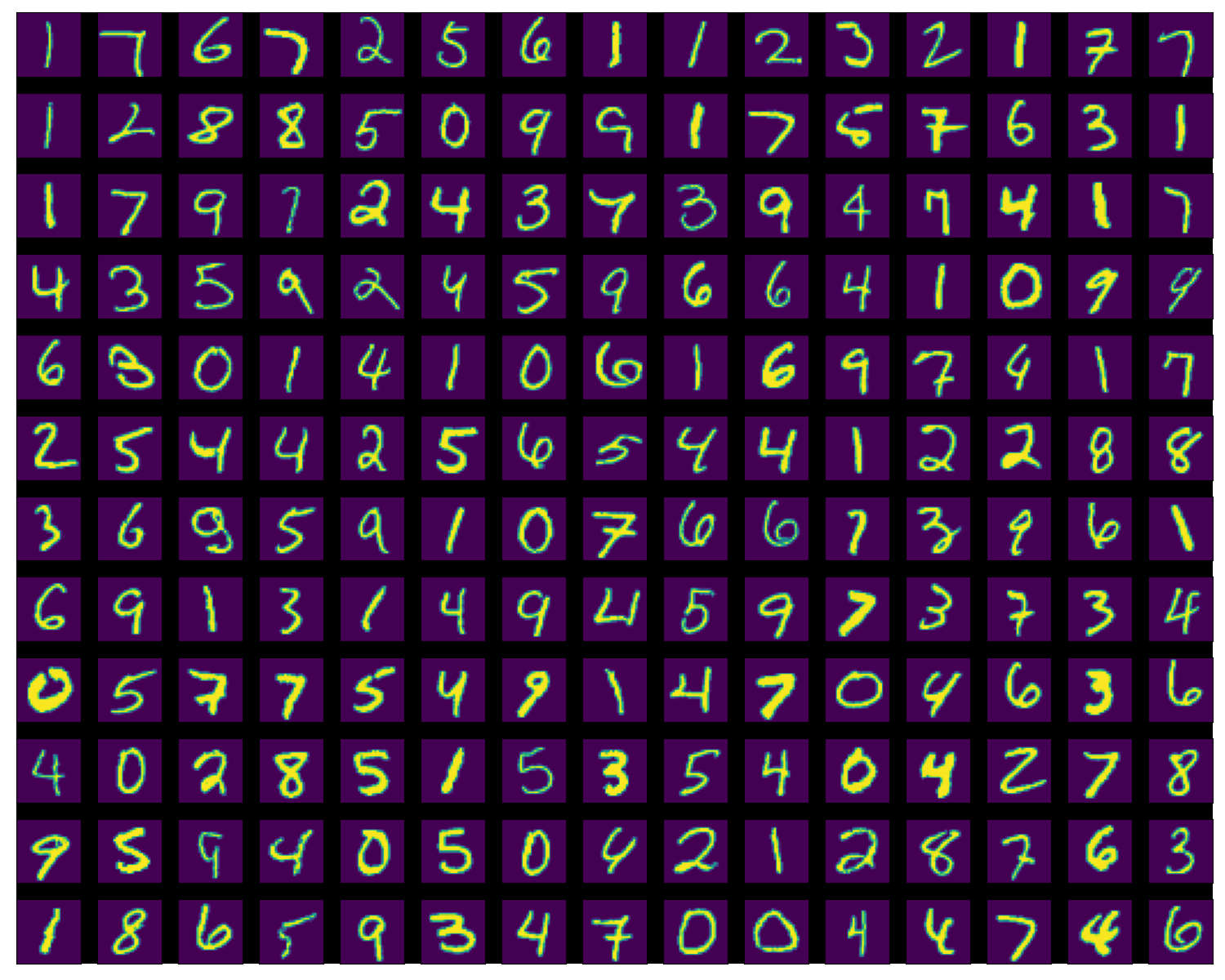}
    \label{fig:mnist_latent_vae_img}}
    \subfigure[CR-VAE on MNIST]{\includegraphics[width=0.48\linewidth]{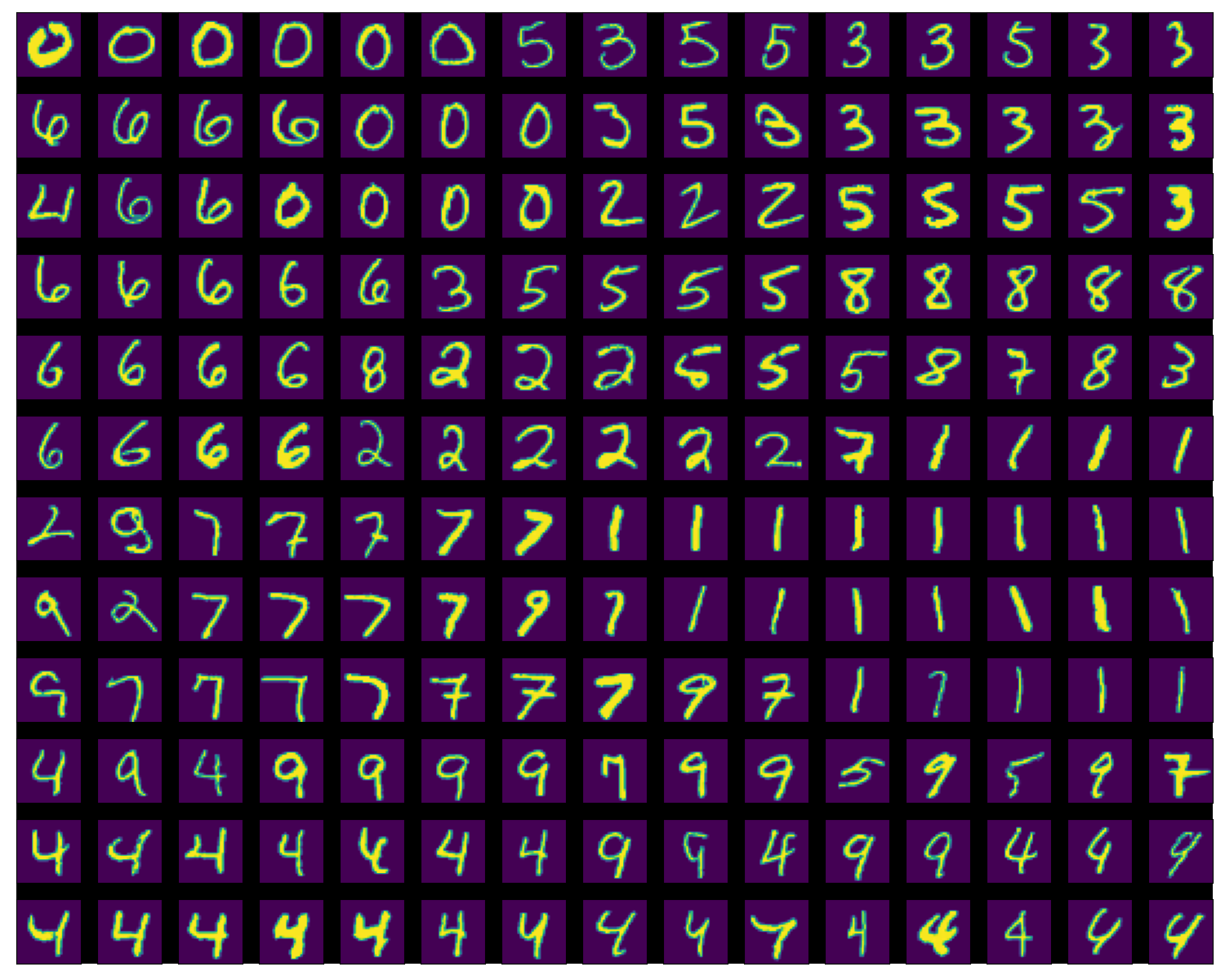}
    \label{fig:mnist_latent_crvae_img}}
    \subfigure[Original VAE on EMNIST]{\includegraphics[width=0.48\linewidth]{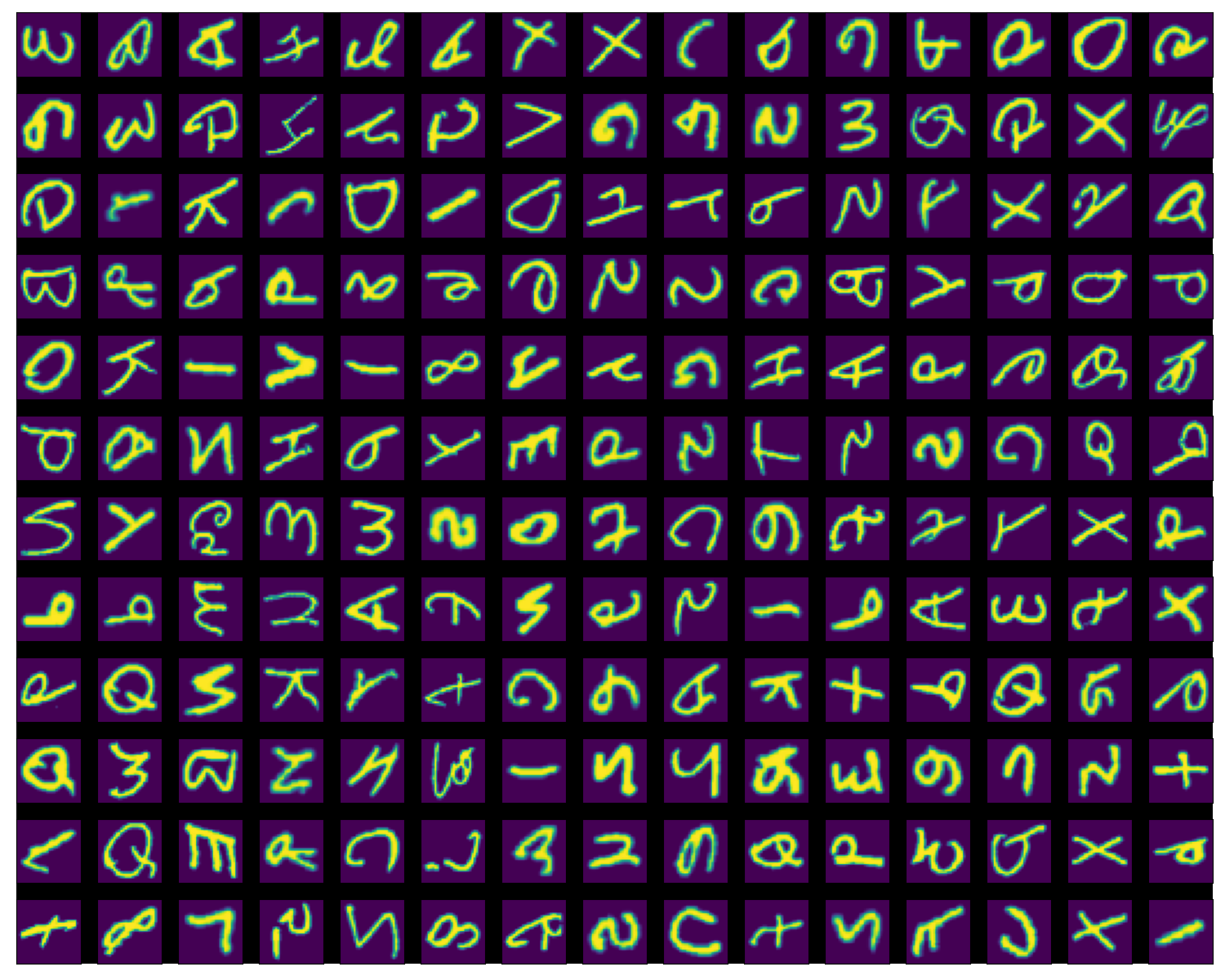}
    \label{fig:emnist_latent_vae_img}}
    \subfigure[CR-VAE on EMNIST]{\includegraphics[width=0.48\linewidth]{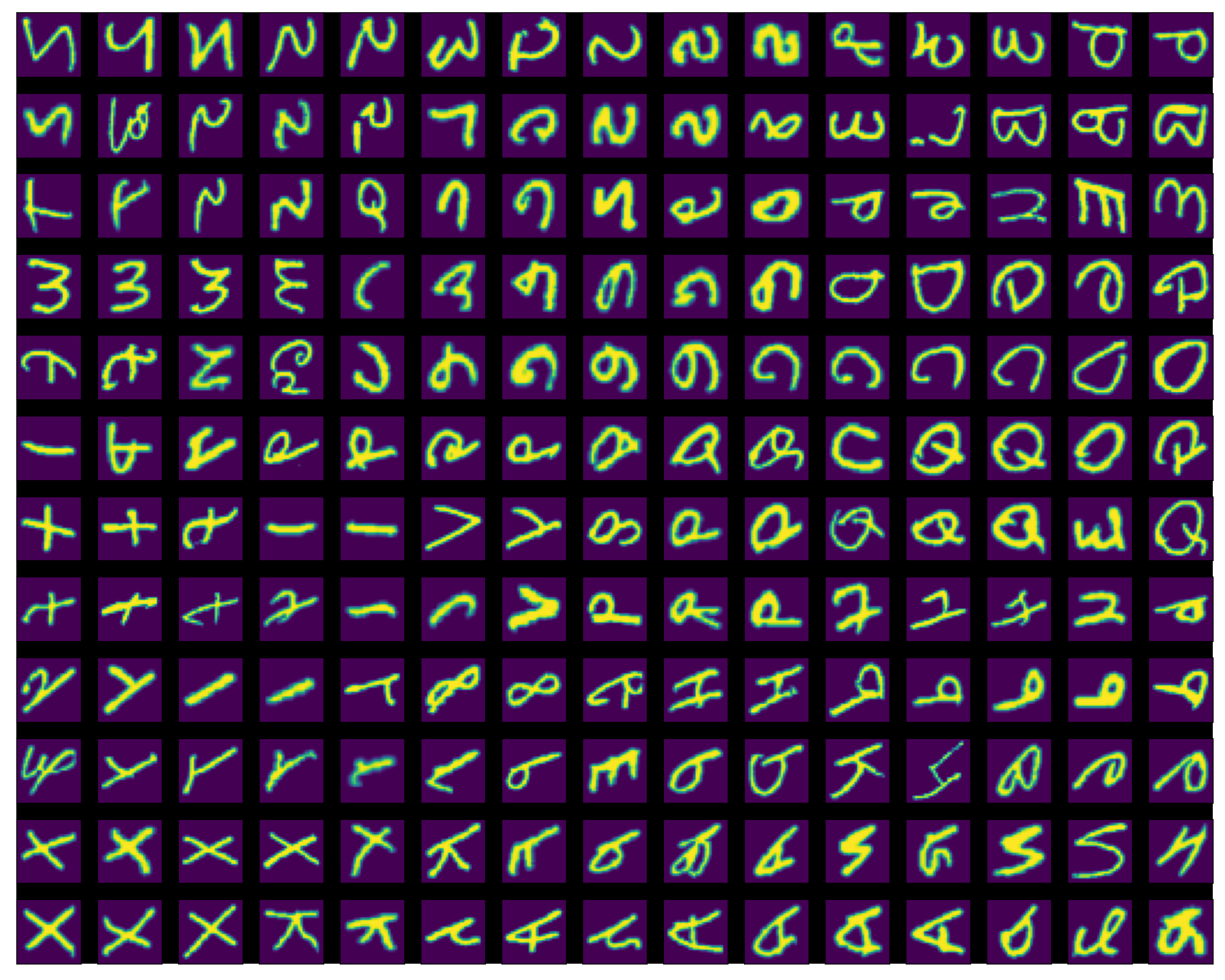}
    \label{fig:emnist_latent_crvae_img}}
    \subfigure[Original VAE on FashionMNIST]{\includegraphics[width=0.48\linewidth]{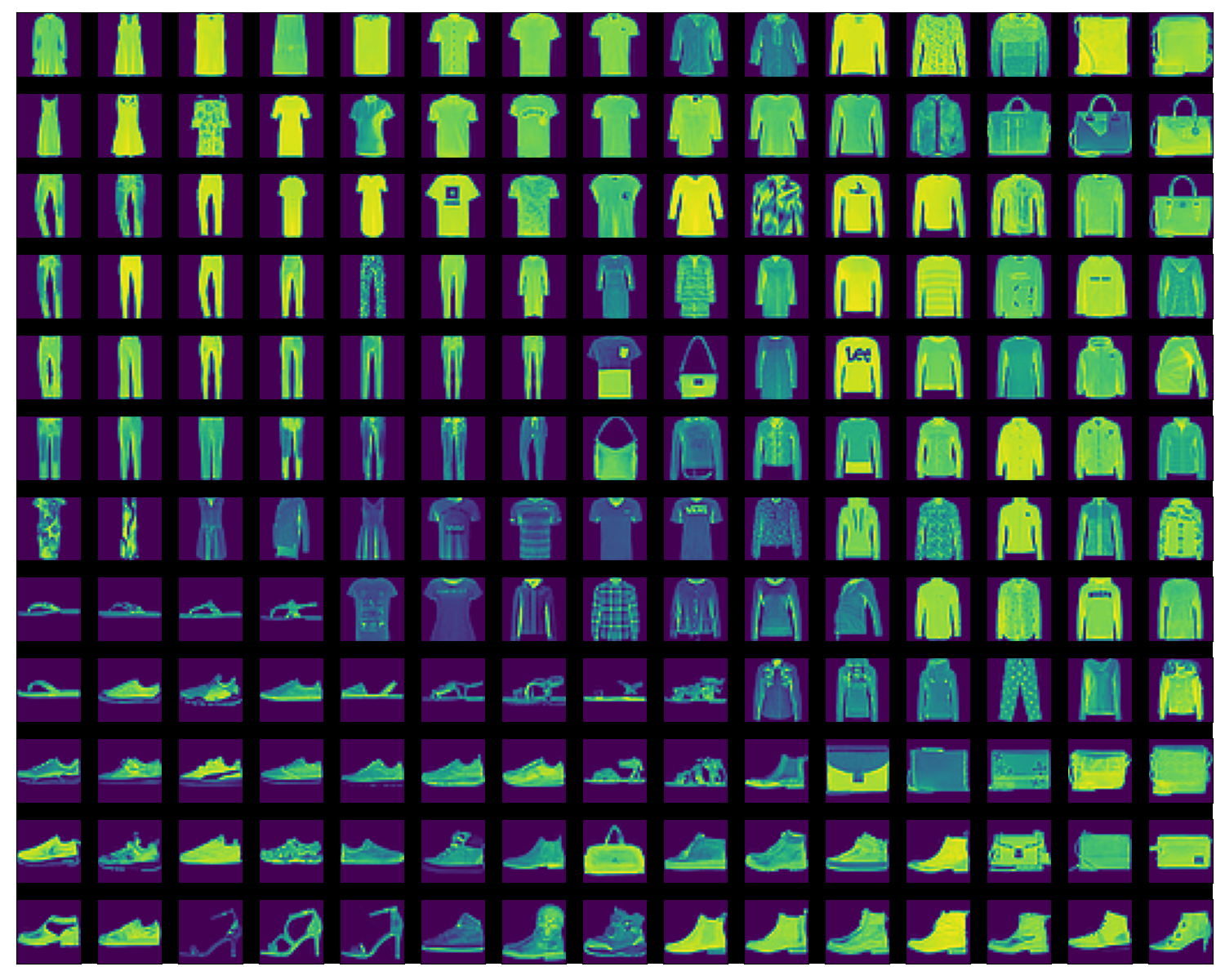}
    \label{fig:fashionmnist_latent_vae_img}}
    \subfigure[CR-VAE on FashionMNIST]{\includegraphics[width=0.48\linewidth]{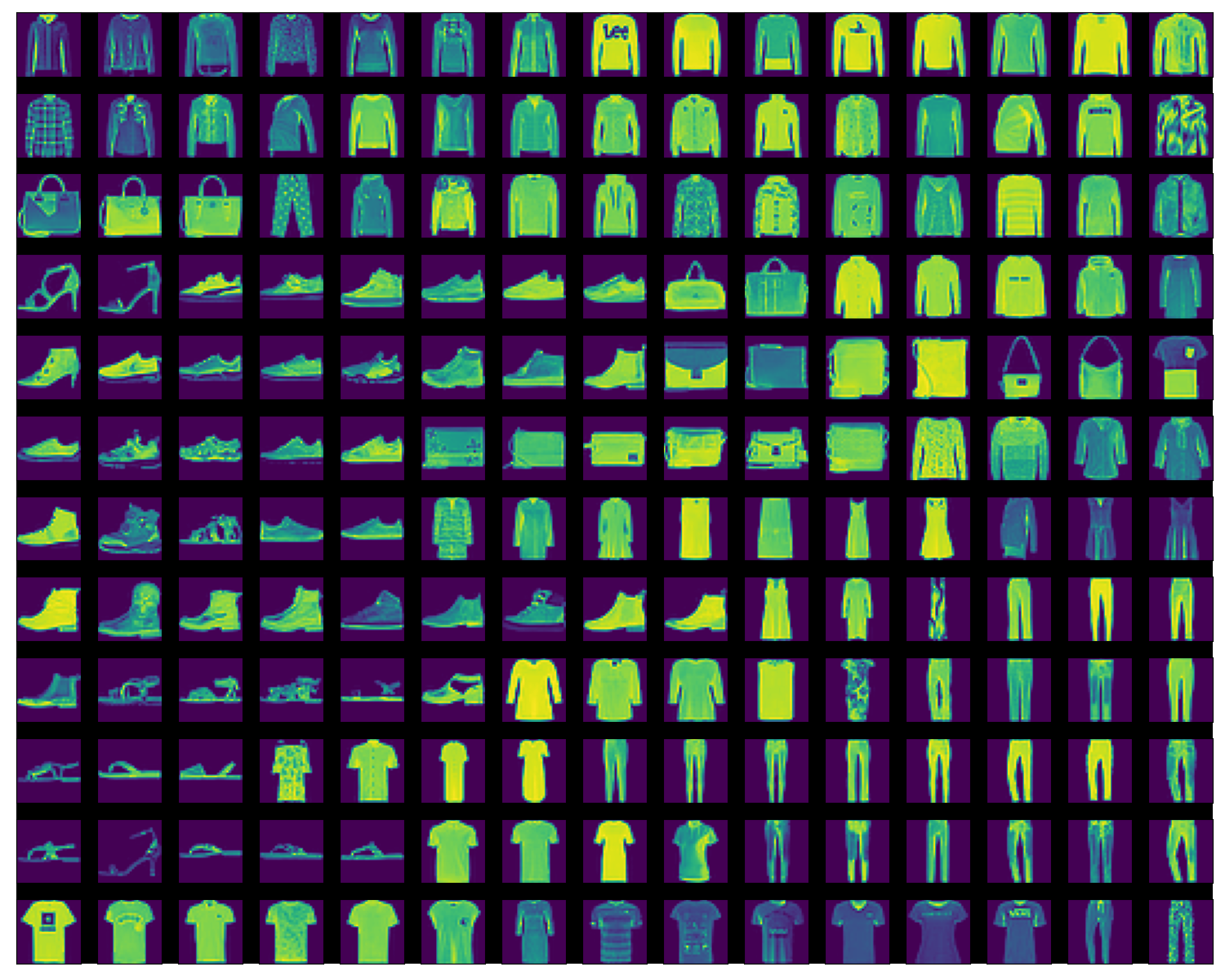}
    \label{fig:fashionmnist_latent_crvae_img}}
    \subfigure[Original VAE on CIFAR10]{\includegraphics[width=0.48\linewidth]{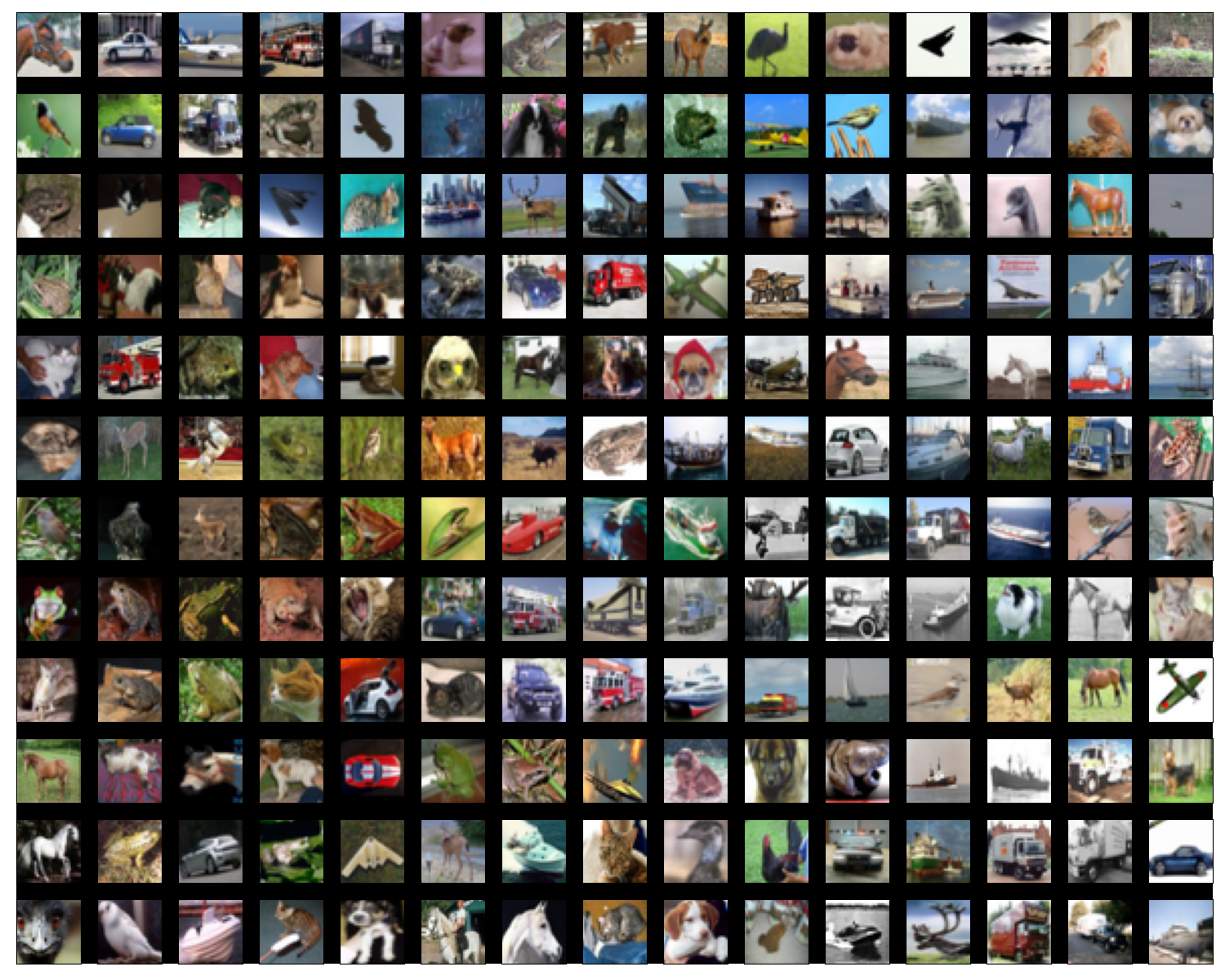}
    \label{fig:cifar_latent_vae_img}}
    \subfigure[CR-VAE on CIFAR10]{\includegraphics[width=0.48\linewidth]{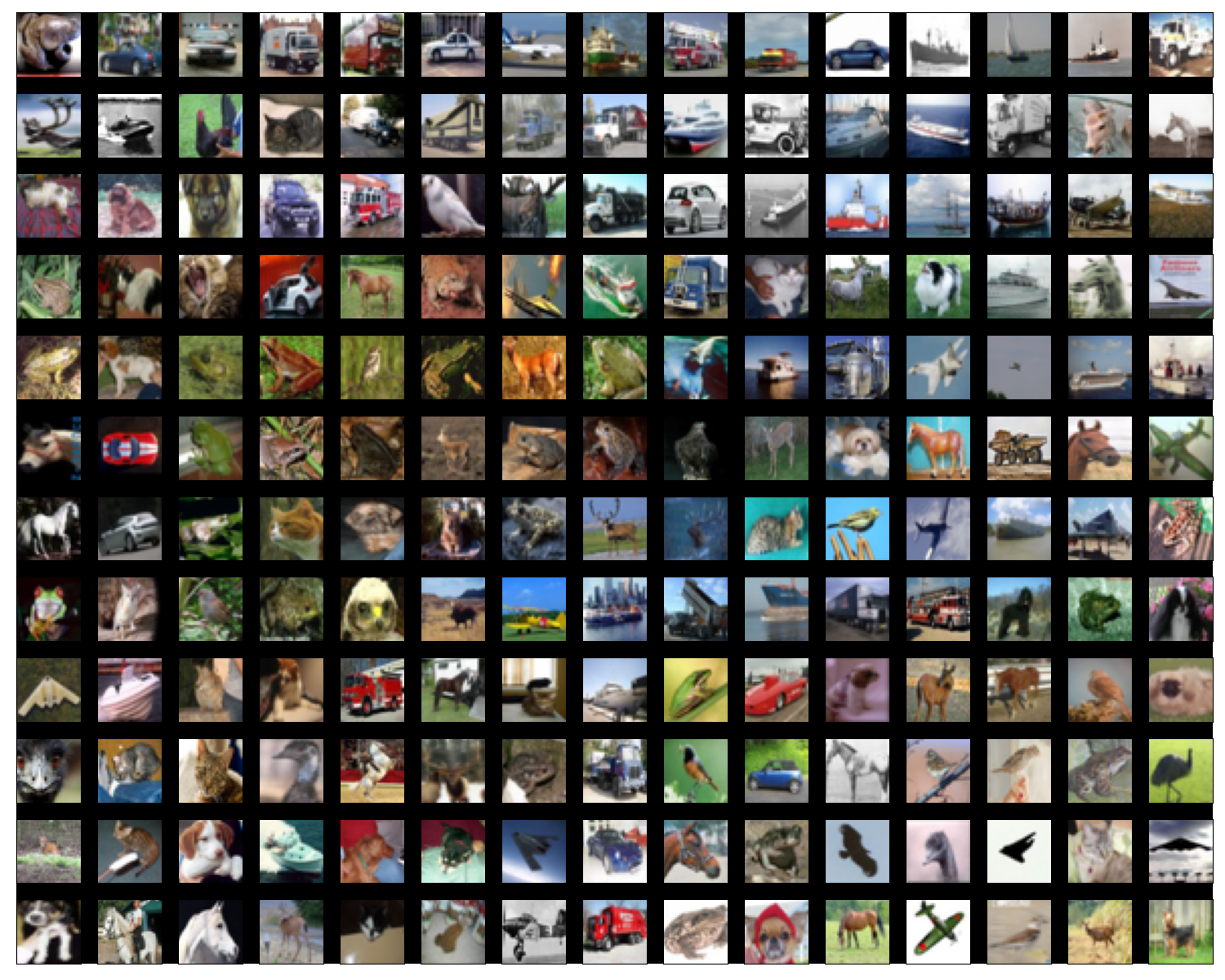}
    \label{fig:cifar_latent_crvae_img}}
    \caption{Image clustering based on their latent representations of VAE and CR-VAE for the MNIST, EMNIST, FashionMNIST and CIFAR10 datasets. We observe similar behavior of the embeddings as in Figure~\ref{fig:latent_2d}. CR-VAE representations (right column) lead to smoother transitions between neighboring images for MNIST and EMNIST. For the FashionMNIST dataset, where posterior collapse is absent for both models, there is a smooth transition between neighbouring images in the grid. For CIFAR10, both models lead to locally cohesive cluster of images, but a universal smoothness of the images in the grid is absent.}
        \label{fig:latent_images}
\end{figure}

We compared the learned latent space structures of the original VAE and CR-VAE, and observed that CR-VAE learns distinct representation modes, which are concentrations of same-colored points, and a structured latent space, as shown in Figure~\ref{fig:mnist_latent_crvae} and~\ref{fig:emnist_latent_crvae}. In contrast, the original VAE suffers from \textit{posterior collapse} and fails to create a distinct representation for each class, as shown in Figure~\ref{fig:mnist_latent_vae} and~\ref{fig:emnist_latent_vae}. To better visualize this effect, we observe Figures~\ref{fig:mnist_latent_vae_img} to~\ref{fig:emnist_latent_crvae_img}, which show image clustering based on their latent representation for each method. CR-VAE representations lead to smoother transitions over neighboring images in the grid of these figures.

Furthermore, the latent representations of FashionMNIST in Figures~\ref{fig:fashionmnist_latent_vae} and~\ref{fig:fashionmnist_latent_crvae} show almost the same structure, although Figure~\ref{fig:fashionmnist_latent_crvae} demonstrates that CR-VAE structures a latent space with more solid clusters. Nevertheless, VAE does not exhibit any major collapse, which is also suppoted by the representation-based image clustering in Figures~\ref{fig:fashionmnist_latent_vae_img} and ~\ref{fig:fashionmnist_latent_crvae_img}.
For the last considered dataset in Figures~\ref{fig:cifar_latent_vae} and~\ref{fig:cifar_latent_crvae}, where NLL is kept relatively high for all models considered (Table~\ref{tab:results2}), both the VAE and CR-VAE cannot achieve distinct modes in their learned latent space, even though we do not observe a complete collapse. This is also supported by Figures~\ref{fig:cifar_latent_vae_img} and~\ref{fig:cifar_latent_crvae_img} where we see a color and and class correlation between neighboring images in the grid (For bigger images please see the Technical Appendix).

\section{Conclusions}
In this work, we present \textit{Contrastive Regularization for VAE (CR-VAE)}, a method that successfully tackles the problem of \textit{posterior collapse} found in VAE. 
Our method combines a contrastive regularization objective with the traditional objective of a VAE. 
CR-VAE leads to lower NLL and higher AU and MI between the input and its learned representation. This means that, it tackles \textit{posterior collapse} more efficiently by increasing the amount of information that it encodes between them. In addition, we show that CR-VAE better avoids the collapsing of its posterior by examining the performance of the learned representations on two semi-supervised classification tasks.
Finally, we show that the augmented objective leads to a learned encoder that maintains distinct modes and a structured latent representation space.
In future work, we will examine our method on bigger and more complex datasets and test the learned representations on more downstream tasks.

\section{Acknowledgements}

This project has received funding from the Deutsche Forschungsgemeinschaft (DFG, German Research Foundation) - No $\#430054590$ (TRAIN).

\bibliographystyle{IEEEtran}
\bibliography{IEEEexample}

\newpage
\appendix
\section{Appendix}
\subsection{The nature of likelihood distribution $q_{\theta}(x|z)$.}
\label{sec:reconstruction_error}
As we mentioned in the paper, CR-VAE assumes an isotropic multivariate Gaussian likelihood, i.e. $q_{\theta}(x|z) \sim \mathcal{N}(f(z), {\sigma}^2I)$, where $f(z)$ is the neural network function of the decoder that reconstructs the input $\hat{x}=f(z)$ from $z$. Consequently, the reconstruction error is calculated as:
\begin{equation} \label{mse_reconstruction}
\begin{split}
	&\log q(x|z) = \log \frac{1}{\sigma\sqrt{2\pi}} + \log \exp (-\frac{1}{2}\frac{(x-f(z))^2}{\sigma^2}) \\
    &\sim \log \exp (-\frac{1}{2}\frac{(x-f(z))^2}{\sigma^2}) = - \frac{1}{2\sigma^2} (x-f(z))^2 \\
    &\sim - (x-f(z))^2
    \;. \nonumber
	\end{split}
\end{equation}
The Mean Square Error (MSE) is thus the reconstruction error we are using for our experiments. This can lead to a lossy compressor ~\cite{Yu2020ATO}. However, due to the contrastive regularization term of CR-VAE this phenomenon was not observed.

Meanwhile, the baseline approaches we compare against on Table 2, assume a Bernoulli distributed likelihood. In these cases, each dimension $i$ of $z$, i.e. $z^i$ is treated as a parameter $p_{out}$ of a Bernoulli distribution. This assumption dictates a Binary Cross Entropy loss to compute their reconstruction error:
\begin{equation} \label{bce_reconstruction}
\begin{split}
	& \log q(x|z) = \log q_{\theta}(x|p_{out}) = \log q_{p_{out}}(X=x) = \\
    & \log (p_{out}^{x}(1-p_{out}^{x})) = x \log p_{out} + (1-x)\log (1- p_{out})
    \;. \nonumber
	\end{split}
\end{equation}


\subsection{Experimental Setup Details}
\label{sec:experiments}

\subsubsection{Datasets.}
We train \textit{CR-VAE} on the MNIST~\cite{mnist}, EMNIST~\cite{emnist}, FashionMnist~\cite{fashionmnist}, CIFAR10~\cite{cifar} and Omniglot~\cite{omniglot} datasets. MNIST consists of 28x28 pixel grey-scale labeled images of handwritten digits with 10 classes and is split into a 60K train set and a 10K test set. We are using the balanced EMNIST dataset that consists of 28x28 pixel grey-scale labeled images of handwritten character digits with 47 balanced classes and is split into a 112.8K train set and a 18.2K test set. FashionMNIST is a dataset on article images with the same characteristics as MNIST. The CIFAR10 dataset is composed of 60K color images with a resolution of 32x32 pixels, which are classified into 10 distinct classes, each containing 6K images. The dataset is divided into a training set, which contains 50K images, and a test set, which contains 10K images.
Finally, Omniglot is a collection of handwritten characters from a variety of different alphabets and scripts, including alphabets used in ancient and modern scripts. The dataset includes 1623 different handwritten characters from 50 different alphabets, with 20 examples of each character. Its images is again grayscale, but they vary in size around 100 pixels in each dimension. In this work, we resize them to the dimension of the datasets above.
%
\subsection{Data Augmentation}
\label{sec:data_aug}
In the context of CR-VAE, we need to augment the data to obtain a positive sample of the input. In this paper we perform the operations of Table 
\ref{tab:augmentation} the input data, with the presented order.
\begin{table}\centering
\caption{Data Augmentation Details}\label{tab:augmentation}
\scriptsize
\begin{tabular}{lrr}\toprule
Random Resized Crop & \\
Random Horizontal flip with probability 0.5 & \\
Color Jitttering with probability 0.8 &brightness: 0.4 \\
&contrast: 0.4 \\
&saturation: 0.4 \\
&hue: 0.1 \\
Random Gray Scale with probability 0.2 &(for CIFAR-10) \\
\bottomrule
\end{tabular}
\end{table}

\subsection{Model Architecture.}
In this section we present the Table~\ref{tab:arch} with the specifications of the model's architecture.

\subsection{Training.}
\label{sec:training_details}
We use stochastic gradient descent as our optimizer with a momentum value of $0.9$, a weight decay value of $10^{-8}$, and an initial learning rate of $10^{-3}$. We use a batch size of $256$ samples and train for $50$ epochs on a single GPU, with the learning rate being reduced when $\mathcal{L}_{CR-VAE}$ is not being reduced for $20$ epochs  by a factor of $0.9$. More details about the hyperparameters can be found in Table~\ref{tab:hyperparameters}.
\begin{table}[!ht]\centering
\caption{Training Hyper-parameters. Parameters inside parenthesis were only used for training on Omniglot to meet the comparison standards with the baselines.}
\label{tab:hyperparameters}
\scriptsize
\begin{tabular}{lrr}\toprule
\multicolumn{2}{c}{Training Hyper-parameters} \\\midrule
epochs &50 \\
batch size &256 (50) \\
learning rate &$10^{-3}$ \\
latent dimensions &16 (32) \\
optimizer & SGD (ADAM) \\
SGD weight decay &$10^{-8}$ \\
KNN voters &200 \\
reconstruction reror & MSE \\
$\gamma$ - MNIST/EMNIST/FashionMNIST/CIFAR10 & $1$ \\
$\gamma$ - Omniglot & $0.4$ \\
\bottomrule
\end{tabular}
\end{table}
\onecolumn
\begin{table}[h]\centering
\caption{The architecture details of VAE. For an input size of $28\times28$, i.e. MNIST, EMNIST and FashionMNIST, the padding for at the decoder layers is [1, 1, 0] from the first to the last layer. The respective output padding values are [0, 0, 1]. For the CIFAR10 dataset, with an image size of $32\times32$ the respective padding and output padding values are [1, 1, 1] and [1, 1, 1]}
\label{tab:arch}
\scriptsize
\begin{tabular}{lllr}\toprule
Layer &Encoder &Decoder \\\midrule
1 &Conv2d(32, kernel=3, stride=2) &Linear(16, 128) \\
&BatchNorm(32) & \\
&ReLU & \\
\midrule
2 &Conv2d(64, kernel=3, stride=2, padding=1) &ConvTranspose2d(256, kernel=3, stride=2, padding, output\_padding) \\
&BatchNorm(64) &BatchNorm(256) \\
&ReLU &ReLU \\
\midrule
3 &Conv2d(128, kernel=3, stride=2, padding=1) &ConvTranspose2d(128, kernel=3, stride=2, padding, output\_padding) \\
&BatchNorm(128) &BatchNorm(128) \\
&ReLU &ReLU \\
\midrule
4 &Linear(128, 16) x2 &ConvTranspose2d(64, kernel=3, stride=2, padding, output\_padding) \\
& &BatchNorm(64) \\
& &ReLU \\
\midrule
5 & &ConvTranspose2d(32, kernel=3, stride=2, padding, output\_padding) \\
& &BatchNorm(32) \\
& &Sigmoid \\
\bottomrule
\end{tabular}
\end{table}

\end{document}